\documentclass[10pt,journal,compsoc]{IEEEtran}
\ifCLASSOPTIONcompsoc
  \usepackage[nocompress]{cite}
\else
  \usepackage{cite}
\fi

\usepackage{graphicx}
\usepackage{amsmath,amssymb} % define this before the line numbering.
\usepackage{color}
\usepackage{dsfont}
\usepackage[algo2e]{algorithm2e}
\usepackage{algorithmic}
\usepackage{algorithm}
\usepackage{subfigure}
\usepackage{booktabs}
\usepackage{multirow}
\usepackage[pagebackref=true,breaklinks=true,letterpaper=true,colorlinks,bookmarks=false,citecolor=blue]{hyperref}
\ifCLASSINFOpdf
\else
\fi
\begin{document}
%
% paper title
% Titles are generally capitalized except for words such as a, an, and, as,
% at, but, by, for, in, nor, of, on, or, the, to and up, which are usually
% not capitalized unless they are the first or last word of the title.
% Linebreaks \\ can be used within to get better formatting as desired.
% Do not put math or special symbols in the title.
%\title{When Sparse Subspace Clustering Meets Neural Networks}
\title{Deep Sparse Subspace Clustering}

% author names and IEEE memberships
% note positions of commas and nonbreaking spaces ( ~ ) LaTeX will not break
% a structure at a ~ so this keeps an author's name from being broken across
% two lines.
% use \thanks{} to gain access to the first footnote area
% a separate \thanks must be used for each paragraph as LaTeX2e's \thanks
% was not built to handle multiple paragraphs
%
%
%\IEEEcompsocitemizethanks is a special \thanks that produces the bulleted
% lists the Computer Society journals use for "first footnote" author
% affiliations. Use \IEEEcompsocthanksitem which works much like \item
% for each affiliation group. When not in compsoc mode,
% \IEEEcompsocitemizethanks becomes like \thanks and
% \IEEEcompsocthanksitem becomes a line break with idention. This
% facilitates dual compilation, although admittedly the differences in the
% desired content of \author between the different types of papers makes a
% one-size-fits-all approach a daunting prospect. For instance, compsoc 
% journal papers have the author affiliations above the "Manuscript
% received ..."  text while in non-compsoc journals this is reversed. Sigh.

\author{Xi~Peng,
           Jiashi~Feng,
        Shijie~Xiao,
           Jiwen~Lu~\IEEEmembership{Senior Member,~IEEE},    Zhang~Yi~\IEEEmembership{Fellow,~IEEE},     Shuicheng~Yan~\IEEEmembership{Fellow,~IEEE},
    % <-this % stops a space
\IEEEcompsocitemizethanks{\IEEEcompsocthanksitem X. Peng and Z. Yi are with College of Computer Science, Sichuan University, Chengdu, 610065, China.\protect\\
% note need leading \protect in front of \\ to get a newline within \thanks as
% \\ is fragile and will error, could use \hfil\break instead.
E-mail: pengx.gm@gmail.com; zhangyi@scu.edu.cn
\IEEEcompsocthanksitem J. Feng is with Department of Electrical and Computer Engineering at National University of Singapore.
\IEEEcompsocthanksitem Shijie~Xiao is with
               OmniVision Technologies Singapore Pte. Ltd., Singapore.
\IEEEcompsocthanksitem J.~Lu is with 
            Department of Automation, Tsinghua University, Beijing 100084, China. 
\IEEEcompsocthanksitem S. Yan is with Department of Electrical and Computer Engineering at National University of Singapore and 360 Artificial Intelligence Institute, China. }% <-this % stops an unwanted space
\thanks{}}

% note the % following the last \IEEEmembership and also \thanks - 
% these prevent an unwanted space from occurring between the last author name
% and the end of the author line. i.e., if you had this:
% 
% \author{....lastname \thanks{...} \thanks{...} }
%                     ^------------^------------^----Do not want these spaces!
%
% a space would be appended to the last name and could cause every name on that
% line to be shifted left slightly. This is one of those "LaTeX things". For
% instance, "\textbf{A} \textbf{B}" will typeset as "A B" not "AB". To get
% "AB" then you have to do: "\textbf{A}\textbf{B}"
% \thanks is no different in this regard, so shield the last } of each \thanks
% that ends a line with a % and do not let a space in before the next \thanks.
% Spaces after \IEEEmembership other than the last one are OK (and needed) as
% you are supposed to have spaces between the names. For what it is worth,
% this is a minor point as most people would not even notice if the said evil
% space somehow managed to creep in.

% The paper headers
\markboth{Journal of \LaTeX\ Class Files,~Vol.~14, No.~8, August~2015}%
{Shell \MakeLowercase{\textit{et al.}}: Bare Demo of IEEEtran.cls for Computer Society Journals}
% The only time the second header will appear is for the odd numbered pages
% after the title page when using the twoside option.
% 
% *** Note that you probably will NOT want to include the author's ***
% *** name in the headers of peer review papers.                   ***
% You can use \ifCLASSOPTIONpeerreview for conditional compilation here if
% you desire.

% The publisher's ID mark at the bottom of the page is less important with
% Computer Society journal papers as those publications place the marks
% outside of the main text columns and, therefore, unlike regular IEEE
% journals, the available text space is not reduced by their presence.
% If you want to put a publisher's ID mark on the page you can do it like
% this:
%\IEEEpubid{0000--0000/00\$00.00~\copyright~2015 IEEE}
% or like this to get the Computer Society new two part style.
%\IEEEpubid{\makebox[\columnwidth]{\hfill 0000--0000/00/\$00.00~\copyright~2015 IEEE}%
%\hspace{\columnsep}\makebox[\columnwidth]{Published by the IEEE Computer Society\hfill}}
% Remember, if you use this you must call \IEEEpubidadjcol in the second
% column for its text to clear the IEEEpubid mark (Computer Society jorunal
% papers don't need this extra clearance.)

% use for special paper notices
%\IEEEspecialpapernotice{(Invited Paper)}

% for Computer Society papers, we must declare the abstract and index terms
% PRIOR to the title within the \IEEEtitleabstractindextext IEEEtran
% command as these need to go into the title area created by \maketitle.
% As a general rule, do not put math, special symbols or citations
% in the abstract or keywords.
\IEEEtitleabstractindextext{%

\begin{abstract}
In this paper, we present a deep extension of Sparse Subspace Clustering, termed Deep Sparse Subspace Clustering (DSSC). Regularized by the unit sphere distribution assumption for the learned deep features, DSSC can infer a new data affinity matrix by simultaneously satisfying the sparsity principle of SSC and the nonlinearity given by neural networks. One of the appealing advantages brought by DSSC is: when original real-world data do not meet the class-specific linear subspace distribution assumption, DSSC can employ neural networks to make the assumption valid with its hierarchical nonlinear transformations. To the best of our knowledge, this is among the first deep learning based subspace clustering methods. Extensive experiments are conducted on four real-world datasets to show the proposed DSSC is significantly superior to 12  existing methods for subspace clustering.
\end{abstract}

% Note that keywords are not normally used for peerreview papers.
}

% make the title area
\maketitle

% To allow for easy dual compilation without having to reenter the
% abstract/keywords data, the \IEEEtitleabstractindextext text will
% not be used in maketitle, but will appear (i.e., to be "transported")
% here as \IEEEdisplaynontitleabstractindextext when the compsoc 
% or transmag modes are not selected <OR> if conference mode is selected 
% - because all conference papers position the abstract like regular
% papers do.
\IEEEdisplaynontitleabstractindextext
% \IEEEdisplaynontitleabstractindextext has no effect when using
% compsoc or transmag under a non-conference mode.

% For peer review papers, you can put extra information on the cover
% page as needed:
% \ifCLASSOPTIONpeerreview
% \begin{center} \bfseries EDICS Category: 3-BBND \end{center}
% \fi
%
% For peerreview papers, this IEEEtran command inserts a page break and
% creates the second title. It will be ignored for other modes.
\IEEEpeerreviewmaketitle

\IEEEraisesectionheading{\section{Introduction}\label{sec:introduction}}

\IEEEPARstart{S}{ubspace}  clustering aims at simultaneously implicitly finding out an underlying subspace to fit each group of data points and performing clustering based on the learned subspaces, which has attracted a lot of interest from the computer vision and image processing community~\cite{Vidal2005}. Most existing subspace clustering methods can be roughly divided into following categories: algebraic methods~\cite{Costeira1998:multibody, Vidal2005}, iterative methods~\cite{Bradley2000,Lu2006Com}, statistical methods~\cite{Ma2007,Rao2008}, and spectral clustering based methods~\cite{Ng2002,Shi2000,Yang:2008:Un,Chen2009,Nie2011}.

Recently, a large number of spectral clustering based methods have been proposed~\cite{Elhamifar2009, Elhamifar2011, Elhamifar2013, Feng2014:Robust, Favaro2011, Hu2014:Smooth, Ji2015:Shape, Liu2013, Liu2010, Lu2012, Vidal2014:Low, Soltanolkotabi2014:Robust, Zhang:2015ku, He2015:Robust, Zhang:2016eq, Lee2016Mini, Liu2012FRR, Cheng2010,You2016:Scale,Yang2016:L0}, which first form an affinity matrix using the linear reconstruction coefficients of the whole data set and then obtain clustering results by applying spectral clustering on the affinity matrix. Those methods differ from each other mainly in their adopted priors on the coefficients. For example, $\ell_1$-norm based sparse subspace clustering (SSC)~\cite{Elhamifar2009,Elhamifar2013} and its $\ell_0$-norm based variant \cite{Yang2016:L0}, low rank representation (LRR)~\cite{Liu2013,Liu2010}, and thresholding ridge regression (TRR)~\cite{Peng2015:Robust,Peng2016:Constructing_abbv} build the affinity matrix using the linear representation coefficients under the constraint of $\ell_1$-, nuclear-, and $\ell_2$-norm, respectively. Formally, SSC, LRR, TRR, as well as many of their variants learn the representation coefficients to build the affinity matrix via:
\begin{equation}
	\label{eq1.1}
	\min_{\mathbf{C}} \mathcal{L}(\mathbf{X}-\mathbf{X}\mathbf{C})+\mathcal{R}(\mathbf{C}),
\end{equation}
where $\mathbf{C}\in \mathds{R}^{n\times n}$ denotes the linear representation of the input $\mathbf{X}\in \mathds{R}^{d\times n}$. Here, $d$ denotes the dimension of data and $n$ is the number of data points. $\mathcal{R}(\mathbf{C})$ denotes certain imposed structure prior over $\mathbf{C}$, and the choice of representation error function $\mathcal{L}(\cdot)$ is usually dependent on the distribution assumption of $\mathbf{X}$, \emph{e.g.} a typical loss function is $\mathcal{L}(\mathbf{X}-\mathbf{X}\mathbf{C})=\|\mathbf{X}-\mathbf{X}\mathbf{C}\|_{F}$.

\begin{figure}[!t]
\centering
\includegraphics[width=0.48\textwidth]{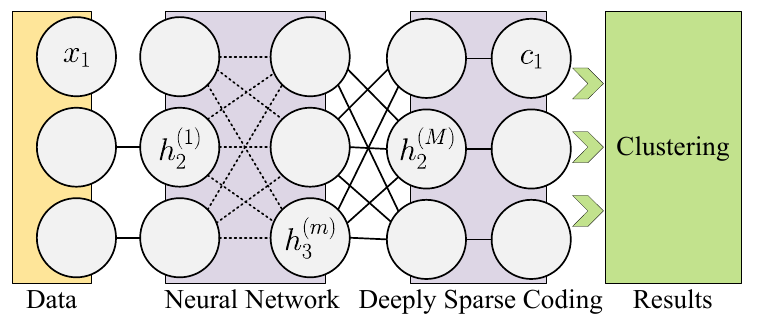}
\caption{\label{fig1} The flowchart of the proposed DSSC method. For a given data set $\mathbf{X}=[\mathbf{x}_{1}, \mathbf{x}_{2}, \cdots, \mathbf{x}_{n}]$, we project them into the feature space as $\mathbf{H}^{(M)}=[\mathbf{h}_{1}^{(M)}, \mathbf{h}_{2}^{(M)}, \cdots, \mathbf{h}_{n}^{(M)}]$ by using a set of hierarchical nonlinear transformations and learn the self sparse representation of input at the top layer of the neural network, where $M$ denotes the top layer of the neural network. Once the neural network converges, we apply spectral clustering on the affinity matrix built by the obtained representation like SSC. Noted that the proposed neural network is based on a novel structure which simultaneously enjoys the sparsity of SSC and the nonlinearity of neural networks.}
\end{figure}

Although those methods have achieved impressive performance for subspace clustering, they generally suffer from the following limitations. First of all, those methods assume that each sample can be  \textit{linearly} reconstructed by the whole sample collection. However, in real-world cases, the data may not be linearly represented by each other in the input space. Therefore, performance of those methods usually drop in practice. To address this problem, several recent works~\cite{Patel2013:Latent,Patel2014:Image,Xiao2015:Robust,Yin2016:Kernel} have developed kernel-based approaches which have shown their effectiveness in subspace clustering. However,  kernel-based approaches are similar to  template-based approaches, whose performance heavily depends on  the choice of kernel functions. Moreover, the approaches cannot give explicit nonlinear transformations, causing difficulties in handling large-scale data sets.

Inspired by the remarkable success of deep learning in various applications~\cite{hinton2012DeepSpeech,Alex2012:ImageNet}, in this work, we propose a new  subspace clustering framework based on neural networks, termed deep sparse subspace clustering (DSSC). The basic idea of DSSC (see Figure~\ref{fig1}) is simple but effective. It uses a neural network to project data into another space in which SSC is valid to the nonlinear subspace case. Unlike most existing subspace clustering methods, our method simultaneously learns \textit{a set of hierarchical transformations} parametrized by a  neural network and \textit{the reconstruction coefficients} to represent each mapped sample as a combination of others. Compared with kernel based approaches, DSSC is a deep instead of shallow model which can  \textit{explicitly} map samples from the input space into a latent space, with parameters in the transformations learned in a data-driven way. To the best of our knowledge, DSSC is the first deep extension of SSC, which satisfies the sparsity principle of SSC and meanwhile makes SSC valid to nonlinear subspace case. 

The contribution of this work is twofold. From the view of subspace clustering, we show how to make it benefit from the success of deep neural networks so that the nonlinear subspace clustering could be achieved. From the view of neural networks, we show that it is feasible to integrate the advantages of existing subspace clustering methods and deep learning to develop new unsupervised learning algorithms. 

\textbf{Notations:} throughout the paper, \textbf{lower-case bold letters} represent column vectors and \textbf{UPPER-CASE BOLD ONES} denote matrices. $\mathbf{A}^T$ denotes the transpose of the matrix $\mathbf{A}$ and $\mathbf{I}$ denotes an identity matrix.

\section{Related Works}

\textbf{Subspace Clustering:} The past decade saw an upsurge of subspace clustering methods with various applications in computer vision, \emph{e.g.} motion segmentation~\cite{Elhamifar2013,Favaro2011,Liu2013,Lu2012,Rao2008,Poling2014A}, face clustering~\cite{Elhamifar2009,Hu2014:Smooth,Ji2015:Shape,Liu2010,Vidal2014:Low}, image processing~\cite{Ding2001:min,Feng2014:Robust,Yang2016:L0}, multi-view analysis~\cite{Zhang:2015ku}, and video analysis~\cite{Xiao2015:Robust}. Particularly, among these works, spectral clustering based methods have achieved state-of-the-art results. The key of these methods is to learn a satisfactory affinity matrix $\mathbf{A}$ in which $\mathbf{A}_{ij}$ denotes the similarity between the $i$-th and the $j$-th sample. Ideally, $\mathbf{A}_{ij}\ne 0$  only if the corresponding data points $\mathbf{x}_{i}$ and $\mathbf{x}_{j}$ are drawn from the same subspace. To this end, some recent works (\emph{e.g.}~SSC~\cite{Elhamifar2009,Elhamifar2013}) assume that any given sample can be linearly reconstructed by other samples in the input space. Based on the self-representation, an affinity matrix (or called similarity graph) can be constructed and fed to spectral clustering algorithms to obtain the final clustering results. In practice, however, high-dimensional data (such as face images) usually resides on the nonlinear manifold. Unfortunately, linear reconstruction assumption may not be satisfied in the original space and in this case the methods may fail to capture the intrinsic nonlinearity of manifold. To address this limitation, the kernel approach is used to first project samples into a high-dimensional feature space in which the representation of the whole data set is computed~\cite{Patel2013:Latent,Patel2014:Image, Xiao2015:Robust,Yin2016:Kernel}. After that, the clustering result is achieved by performing traditional subspace clustering methods in the kernel space. However, the kernel-based methods behave like template-based approaches which usually require the prior knowledge on the data distribution to choose a desirable kernel function. Clearly, such a prior is hard to obtain in practice. Moreover, they cannot learn an explicit nonlinear mapping functions from data set, thus suffering from the scalability issue and the out-of-sample problem~\cite{Peng2013SSSC,Peng2015SRSC}.

Unlike these classical subspace clustering approaches, our method learns a set of \emph{explicit nonlinear mapping functions} from data set to map the input into another space, and calculates the affinity matrix using the representation of the samples in the new space. 

\textbf{Deep Learning:} Aimed at learning high-level features from inputs, deep learning has shown promising results in numerous computer vision tasks in the scenario of supervised learning, such as image classification~\cite{Alex2012:ImageNet}. In contrast, less attention~\cite{Bengio2013:Rep,Lee2011:Un,Wang2016:Learn} has been paid to the applications with unsupervised learning scheme. Recently, some works~\cite{Huang2014:Deep,Xie2016:Un,Peng2016:Deep_abbv,Wang2015:Learning,Yang2016:Joint,Chen2017:Sub,Ji2017:Deep} have devoted to combining deep learning and unsupervised clustering and shown impressive results over the traditional clustering approaches. These methods share the same basic idea, \textit{i.e.}, using deep learning to learn a good representation and then achieving  clustering with existing clustering methods like k-means. The major differences among them reside on  the neural network structure and the objective function.

Different from these works, our framework is based on a new neural network instead of an existing network. Moreover, our method focuses on subspace clustering rather than clustering, which simultaneously learns the high-level features from inputs and the self-representation in a joint way, whereas these existing methods do not enjoy the effectiveness of the self-expressive subspace clustering. We believe that such a general framework is complementary to existing shallow subspace clustering methods, since it can adopt the loss functions and regularizations in these methods. To the best of our knowledge, this is one of the first several deep subspace clustering methods. It should be pointed out that, our model is also significantly different from \cite{Peng2016:Deep_abbv} as below: 1) \cite{Peng2016:Deep_abbv} performs like manifold learning, which requires the data could be linearly reconstructed in the input space and embeds the obtained sparse representation from input space into latent space. In contrast, our model aims to solve the problem of nonlinear subspace clustering, \textit{i.e.} the data cannot  linearly represented in the input space. 2) In \cite{Peng2016:Deep_abbv}, sparse representation is used as a type of priori, which keeps unchanged. In contrast, this work dynamically seeks an good sparse representation to jointly optimize our neural network. 3) The proposed method can be regarded as a deep nonlinear extension of the well-known SSC, which makes SSC handling nonlinear subspace clustering possible.

\section{Deep Sparse Subspace Clustering}

In this section, we first briefly review SSC, and then present the details of our deep subspace clustering method.

\subsection{Sparse Subspace Clustering}

For a given data set $\mathbf{X}=[\mathbf{x}_{1}, \mathbf{x}_{2}, \cdots,\mathbf{x}_{n}]\in \mathds{R}^{d\times n}$, SSC seeks to linearly reconstruct the $i$-th sample $\mathbf{x}_{i}$ using a few of other samples. In other words, the representation coefficients are expected to be sparse. To achieve this end, the problem is formulated as below, 
\begin{equation}
	\label{eq3.2}
%	\begin{aligned}
		\min_{\mathbf{c}_{i}}
        \hspace{1mm} \frac{1}{2}\|\mathbf{x}_{i}-\mathbf{X}\mathbf{c}_{i}\|_{F}^{2}+\gamma\|\mathbf{c}_{i}\|_{1} \hspace{6mm}
		\mathrm{s.t.}\hspace{1mm}{c}_{ii}=0
%	\end{aligned},
\end{equation}
where $\|\cdot\|_{1}$ denotes $\ell_1$-norm (\emph{i.e.}, the sum of absolute values of all elements in a vector) that acts as a relaxation of  $\ell_0$-norm, and ${c}_{ii}$ denotes the $i$-th element in $\mathbf{c}_{i}$. Specifically, penalizing $\|\mathbf{c}_{i}\|_{1}$ encourages $\mathbf{c}_{i}$ to be sparse, and enforcing the constraint ${{c}_{ii}}=0$ to avoid trivial solutions. To deal with the optimization problem in (\ref{eq3.2}), the alternating direction method of multipliers (ADMM)~\cite{Boyd2011:Dis,Lin2011} is often used.

Once the sparse representation of the whole data set is obtained by solving (\ref{eq3.2}), an affinity matrix in SSC is calculated as $\mathbf{A}=|\mathbf{C}|+|\mathbf{C}|^{T}$, based on which spectral clustering is applied to give clustering results.

\subsection{Deep Subspace Clustering}

In most existing subspace clustering methods including SSC, each sample is encoded as a \emph{linear} combination of the whole data set. However, when dealing with high-dimensional data which usually lie on \emph{nonlinear} manifolds, such methods may fail to  capture the nonlinear structure, thus leading to inferior results. To address this issue, we propose a deep learning based method which maps given samples using \emph{explicit hierarchical transformations} in a neural network, and simultaneously learns the \textit{reconstruction coefficients} to represent each mapped sample as a combination of others.

As shown in Figure~\ref{fig1}, the neural network in our proposed framework consists of $M+1$ stacked layers with $M$ nonlinear transformations, which takes a given sample $\mathbf{x}$ as the input to the first layer. For ease presentation, we make several  definitions below. For the first layer of our neural network, we define its input as $\mathbf{h}^{(0)}=\mathbf{x}\in\mathds{R}^{d}$. Moreover, for the subsequent layers, let 
\begin{equation}
	\label{eq3.new1}
\mathbf{h}^{(m)}=g(\mathbf{W}^{(m)}\mathbf{h}^{(m-1)}+\mathbf{b}^{(m)})\in \mathds{R}^{d^{(m)}}
\end{equation}
be the output of the $m$-th layer (in which $m=1,2,\cdots,M$ indexes the layer), where $g(\cdot)$ is a nonlinear activation function, $d^{(m)}$ is the dimension of the output of the $m$-th layer, $\mathbf{W}^{(m)}\in \mathds{R}^{d^{(m)}\times d^{(m-1)}}$ and $\mathbf{b}^{(m)}\in \mathds{R}^{d^{(m)}}$ denote the weights and bias associated with the $m$-th layer, respectively. In particular, given $\mathbf{x}$ as the input of the first layer, the output at the top layer of our neural network is
\begin{equation}
	\label{eq3.4}
\mathbf{h}^{(M)}  = g(\mathbf{W}^{(M)}\mathbf{h}^{(M-1)}+\mathbf{b}^{(M)}).
\end{equation}

In fact, if denoting the expression above as $f(\mathbf{x})$, we can observe that $f(\cdot): \mathds{R}^{d}\rightarrow \mathbf{R}^{d^{(m)}}$ is a nonlinear function determined by the weights and biases of our neural network (\textit{i.e.}, $\{\mathbf{W}^{(m)},\mathbf{b}^{(m)}\}_{m=1}^{M}$) as well as the choice of  activation function $g(\cdot)$. Furthermore, for $n$ samples, we define $\mathbf{H}^{(M)}$ as the collection of the corresponding outputs given by our neural network, \emph{i.e.}
\begin{equation}
\label{eq:LastLayerOutput}
\mathbf{H}^{(M)}=[\mathbf{h}_{1}^{(M)},\mathbf{h}_{2}^{(M)},\cdots,\mathbf{h}_{n}^{(M)}].
\end{equation}

With the above definitions, we present the objective function of our method in the following form:
\begin{equation}
\label{eq:OurGeneral}
\begin{aligned}
\min_{\{\mathbf{W}^{(m)},\mathbf{b}^{(m)}\}_{m=1}^{M}, \mathbf{C}} 
\mathcal{J}&\hspace{1mm}=\mathcal{J}_{1}+\lambda\mathcal{J}_{2}.
	\end{aligned}
\end{equation}
where $\lambda$ is a positive trade-off parameter, and $\{\mathcal{J}_{i}\}_{i=1}^2$ are defined below. Intuitively, the first term $\mathcal{J}_{1}$ is designed to minimize the discrepancy between $\mathbf{H}^{(M)}$ and its self-expressed representation. Moreover, it  meanwhile regularizes $\mathbf{C}$ for some desired properties. To be specific, $\mathcal{J}_{1}$ can be expressed in the form of 
\begin{equation}
	\label{eq:J1}
	\mathcal{J}_{1} = \mathcal{L}(\mathbf{H}^{(M)}-\mathbf{H}^{(M)}\mathbf{C})+\mathcal{R}(\mathbf{C}) + \mathcal{F}(\mathbf{C}),
\end{equation}
where $\mathcal{F}(\mathbf{C})$ takes the value of $+\infty$ if $\mathbf{C}$ is not in some feasible domains, and $0$ otherwise. Note that, the form of $\mathcal{L}(\cdot), \mathcal{R}(\cdot)$, and $\mathcal{F}(\cdot)$ may be adopted from many existing subspace clustering works. In this paper, we aim to develop a deep extension of SSC and thus take $\mathcal{L}(\cdot)=\|\cdot\|_F^2$, $\mathcal{R}(\cdot)=\|\cdot\|_1$, $\mathcal{F}(\mathbf{C}) = +\infty$ if $diag(\mathbf{C})=\mathbf{0}$ is violated, and $\mathcal{F}(\mathbf{C}) =0$ otherwise. 

The second part $\mathcal{J}_{2}$ is designed to remove an arbitrary scaling factor in the latent space. In this work, we set 
\begin{equation}
	\label{eq:J2}
	\mathcal{J}_{2}=\frac{1}{4}\sum_{i=1}^{n}\|(\mathbf{h}_{i}^{(M)})^{T}\mathbf{h}_{i}^{(M)}-1\|_{2}^{2},
\end{equation}
Noticed that, without the above term, our neural network may collapse in the trivial solutions like $\mathbf{H}^{(M)}=\mathbf{0}$.

With $\{\mathcal{J}_{i}\}_{i=1}^2$ detailed above, the optimization problem of our proposed DSSC can be expressed as follows: %
\begin{align}
\label{eq3.5}
\min_{\mathbf{\Theta},\mathbf{C}}
	&\frac{1}{2}\|\mathbf{H}^{(M)}-\mathbf{H}^{(M)}\mathbf{C}\|_{F}^{2}
	+\gamma\|\mathbf{C}\|_{1}\notag\\ 
	&\hspace{2cm}+\frac{\lambda}{4}\sum_{i=1}^{n}\|(\mathbf{h}_{i}^{(M)})^{T}\mathbf{h}_{i}^{(M)}-1\|_{2}^{2}\notag\\
	\mathrm{s.t.}&\hspace{1mm} diag(\mathbf{C})=0,
\end{align}
where $\mathbf{\Theta}$ denotes the parametric  neural network, \textit{i.e.}, $\mathbf{\Theta}=\{\mathbf{W}^{(m)},\mathbf{b}^{(m)}\}_{m=1}^{M}$. 

\subsection{Optimization}
For ease of presentation, we first rewrite $(\ref{eq3.5})$ as follows:
\begin{align}
\label{eq3.6}
\min_{\mathbf{\Theta},\mathbf{c}_{i}}
	\sum_{i=1}^{n} \Big(\frac{1}{2}\|\mathbf{h}^{(M)}_{i}&-\mathbf{H}_{i}^{(M)}\mathbf{c}_{i}\|_{F}^{2}
	+\gamma\|\mathbf{c}_{i}\|_{1}\notag\\
	&+\frac{\lambda}{4}\|(\mathbf{h}_{i}^{(M)})^{T}\mathbf{h}_{i}^{(M)}-1\|_{2}^{2}\Big),
\end{align}
where $\mathbf{H}_{i}^{(M)}$ is a variant of $\mathbf{H}^{(M)}$, which is obtained by simply replacing $\mathbf{h}_{i}^{(M)}$ in $\mathbf{H}^{(M)}$ with $\mathbf{0}$.

Given $n$ data points, DSSC simultaneously learns $M$ nonlinear mapping functions $\{\mathbf{W}^{(m)}, \mathbf{b}^{(m)}\}_{m=1}^{M}$ and $n$ sparse codes $\{\mathbf{c}_{i}\}_{i=1}^{n}$ by solving (\ref{eq3.6}). As  (\ref{eq3.6}) is a multiple-variable  optimization problem, we employ an alternating minimization algorithm by alternatively updating one of variables while fixing the others.

\textbf{Step 1:} Fix $\mathbf{c}_{i}$ and $\mathbf{H}_{i}^{(m)}$, update $\mathbf{\Theta}$,  (\ref{eq3.6}) can be rewritten as
\begin{equation}	
	\label{eq3.7}
		\min_{\mathbf{\Theta}}
		 \frac{1}{2}\|\mathbf{h}^{(M)}_{i}-\mathbf{H}_{i}^{(M)}\mathbf{c}_{i}\|_{2}^{2}
		 +\alpha_{i}
		 +\frac{\lambda}{4}\|(\mathbf{h}_{i}^{(M)})^{T}\mathbf{h}_{i}^{(M)}-1\|_{2}^{2},
\end{equation}
where $\alpha_{i}=\sum_{j\ne i}\|\mathbf{h}_{j}^{(M)}-\mathbf{H}_{j}^{(M)}\mathbf{c}_{j}\|_{2}^{2}+\lambda\|(\mathbf{h}_{i}^{(M)})^{T}\mathbf{h}_{i}^{(M)}-1\|_{2}^{2}$ is a constant.

To solve (\ref{eq3.7}), we adopt the stochastic sub-gradient descent (SGD) algorithm to obtain the parameters $\{\mathbf{W}^{(m)}$, $\mathbf{b}^{(m)}\}_{m=1}^{M}$. Moreover, we also enforce $\ell_2$-norm on the parameters to avoid overfitting~\cite{Alex2012:ImageNet,nntricks:2012}, where the regularization parameter is fixed as $\varphi=10^{-3}$ in all experiments. Noticed that, (\ref{eq3.7}) could also be solved with mini-batch SGD, especially, when the data size is large. However, the mini-batch SGD may give two issues. First, it introduces a new hyper parameter (\textit{i.e.}, batch size), which increases human effort for model selection. Second, the efficiency may be at the cost of performance degradation \cite{Goyal2017:Acc}.

\textbf{Step 2:} Fix $\{\mathbf{h}_{i}^{(M)}\}_{i=1}^{n}$ and update $\mathbf{c}_{i}$ by
\begin{equation}
	\label{eq3.15}
	 \min_{\mathbf{c}_{i}}\frac{1}{2}\|\mathbf{h}_{i}^{(M)}-\mathbf{H}_{i}^{(M)}\mathbf{c}_{i}\|_{F}^{2}+\gamma\|\mathbf{c}_{i}\|_{1}+\beta_{i},
\end{equation}
where $\beta_{i}=\sum_{j\ne i}\left(\frac{1}{2}\|\mathbf{h}_{j}^{(M)}-\mathbf{H}_{j}^{(M)}\mathbf{c}_{i}\|_{2}^{2}+\gamma\|\mathbf{c}_{j}\|_{1}\right)$ is a constant. Note that, (\ref{eq3.15}) is a standard $\ell_1$-minimization problem faced in SSC, which can be solved by using many existing $\ell_1$-solvers~\cite{Yang2010}. 

\textbf{Step 1} and \textbf{Step 2} are repeated until convergence. After obtaining $\mathbf{C}$, we construct a similarity graph via $\mathbf{A}=|\mathbf{C}|+|\mathbf{C}|^{T}$ and obtain the clustering results based on $\mathbf{A}$. The optimization procedure of DSSC is summarized in Algorithm 1.

\begin{algorithm}[!h]
\label{alg1}
\begin{footnotesize}
\caption{Deep Sparse Subspace Clustering}
\textbf{Input}: A given data set $\mathbf{X}$ and the tradeoff parameters  $\lambda$.\\
%\textbf{Output}: $\{\mathbf{W}^{(m)}, \mathbf{b}^{(m)}\}_{m=1}^{M}$ and $\mathbf{C}$.
// Initialization:\\
Initialize $\{\mathbf{W}^{(m)}, \mathbf{b}^{(m)}\}_{m=1}^{M}$, and $\mathbf{H}^{(0)}=\mathbf{X}$.\\
\For{$m=1, 2\cdots, M$}{
Do forward propagation to get $\{\mathbf{H}^{(m)}\}_{m=1}^{M}$ and $\mathbf{C}$ via solving (\ref{eq3.new1}) and (\ref{eq3.15}), respectively.
}

// Optimization\\
\While{not converge}{
\For{$i=1,2,\cdots,n$}{
Randomly select a data point $\mathbf{x}_{i}$ and let $\mathbf{h}_{i}^{0}=\mathbf{x}_{i}$,\\
%// Forward propagation\\
%\If{$i>1$}{
\For{$m=1, 2\cdots,M$}{
Compute $\mathbf{h}^{(m)}_{i}$ via (\ref{eq3.new1}).
}

Compute $\mathbf{c}_{i}$ using $\mathbf{h}_{i}^{(m)}$ via (\ref{eq3.new1}).
%}

%// Compute gradients\\
\For{$m=M, M-1\cdots,1$}{
Calculate the gradient using the SGD algorithm.
	}
%// Update neural network\\
\For{$m=1,2,\cdots,M$}{ 
Update $\mathbf{W}^{(m)}$ and $\mathbf{b}^{(m)}$ with the gradient.}
}
%// Convergence\\
}
\textbf{Output:} $\{\mathbf{W}^{(m)},\mathbf{b}^{(m)}\}_{m=1}^{M}$ and $\mathbf{C}$.
\end{footnotesize}
\end{algorithm}

\subsection{Discussions}

Our approach DSSC can provide satisfactory subspace clustering performance befitting from following factors. First, different from SSC,  DSSC performs sparse coding in a deep latent space instead of the original one and the latent space is automatically learned in a data-driven manner. After mapping input data into the latent space via the learned transformation matrices, the transformed samples are more favorable for linear reconstruction. Second, DSSC can also be deemed as a deep kernel method which automatically learns transformations in a data-driven way. Considering the demonstrated effectiveness of kernel-based subspace clustering approaches such as ~\cite{Patel2014:Image,Xiao2015:Robust}, DSSC is well-expected to show even better performance for subspace clustering thanks to the representative capacity of deep neural network. 

It should be pointed out that the proposed DSSC adopts similar neural network structure with deep metric learning networks (DMLNs)~\cite{Hu2014:Dis,Schroff2015:Facenet,Wang2014:LFI,Yi2014:DMLPR}, \textit{i.e.}, a set of fully connected layers to perform nonlinear transformation and then perform specific task on the output of neural network. The major differences among them are: 1) the objective functions are different. Our method aims to segment different sample into different subspaces, whereas these metric learning networks aim to learn similarity function that measures how similar or related two data points are; 2) our DSSC is unsupervised, whereas DMLNs are supervised approaches which require the label information to train neural networks. 

%\subsection{Implementation Details}\label{sec:ImpDetail}

% which is defined as follows:
%\begin{equation}
%	\label{eq3.16}
%	g(z)=tanh(z)=\frac{1-e^{-2z}}{1+e^{-2z}},
%\end{equation}
%and the corresponding derivative is calculated as
%\begin{equation}
%	\label{eq3.17}
%	g^{\prime}(z)=tanh^{\prime}(z)=1-tanh^{2}(z).
%\end{equation}

\subsection{Implementation Details}\label{sec:ImpDetail}
In this section, we introduce the implementation details of the used activation functions and the initialization of $\{\mathbf{W}^{(m)}, \mathbf{b}^{m}\}$.

The activation functions can be chosen from various forms. In our experiments, we use the $tanh$ function which is defined as follows:
\begin{equation}
	\label{eq3.16}
	g(z)=tanh(z)=\frac{1-e^{-2z}}{1+e^{-2z}},
\end{equation}
and the corresponding derivative is calculated as
\begin{equation}
	\label{eq3.17}
	g^{\prime}(z)=tanh^{\prime}(z)=1-tanh^{2}(z).
\end{equation}

Regarding the initializations of $\{\mathbf{W}^{(m)}, \mathbf{b}^{m}\}$, we initialize $\mathbf{W}^{(m)}$ as a rectangular matrix with ones at the main diagonal and zeros as other elements. Moreover, $\mathbf{b}^{(m)}$ is initialized as $\mathbf{0}$. Note that, the used networks could also be initialized with an auto-encoder.

\section{Experiments}

In this section, we compare our method with 12 popular subspace clustering methods on four different real-world data sets in terms of four clustering performance metrics.

\subsection{Datasets and Experimental Settings}

\textbf{Data sets:} Four different data sets are used in our experiments, \emph{i.e.} COIL20 object images~\cite{Dataset:COIL20}, the MNIST handwritten digital database~\cite{Dataset:MNIST}, AR facial images~\cite{Dataset:AR}, and the BF0502 video face data set~\cite{Dataset:BF05}.

The COIL20 database contains 1,440 samples distributed over 20 objects, where each image is with the size of $32\times32$. The MNIST data set includes 60,000 handwritten digit images of which the first 2,000 training images and the first 2,000 testing images are used in our experiments, where the size of each image is $28\times 28$.

 The AR database is one of the most popular facial image data sets for subspace clustering. In our experiments, we use a widely-used subset of the AR database~\cite{Zhang2011} which consists of 1,400 undisguised faces evenly distributed over 50 males and 50 females, where the size of each image is $165\times 120$.

The BF0502 data set contains facial images detected from the TV series ``Buffy the Vampire Slayer''. Following~\cite{Xiao2015:Robust}, a subset of BF0502 is used, which includes 17,337 faces in 229 tracks from 6 main casts. Each facial image is represented as a 1,937-dimensional vectors extracted from 13 facial landmark points (\textit{e.g.}, the left and right corners of each eye). In our experiments, we use the first 200 samples from each category, thus resulting in 1,200 images in total.

For the purpose of nonlinear subspace clustering, we use the following four types of features instead of raw data from the COIL20, MNIST, and AR data sets in experiments, \textit{i.e.} dense scale-invariant feature transform (DSIFT)~\cite{Toolkit:SIFT}, the histogram of oriented gradients (HOG)~\cite{Toolkit:HOG}, local binary pattern (LBP)~\cite{Toolkit:LBP}, and local phase quantization (LPQ)~\cite{Toolkit:LPQ}. The details of extracting these features are introduced as follows:
\begin{itemize}
  \item DSIFT: We divide each image into multiple non-overlapping patches and then densely sample SIFT descriptors from each patch. The patch sizes of AR, COIL20, and MNIST are set as $15\times 15$, $8\times 8$, and $4\times 4$, respectively. By concatenating these SIFT descriptors extracted from each image, we obtain a feature vector with the dimension of 11,264 (AR), 2,048 (COIL20), and  6,272 (MNIST).
  \item HOG: We first divide each image into multiple blocks with two scales, \textit{i.e.} $8\times 8$ and $4\times 4$ for AR, and $4\times 4$ and $2\times 2$ for MNIST and COIL20. Then, we extract a 9-dimensional HOG feature from each block. By concatenating these features for each image, the dimension of the  feature vector are 13,770 (AR), 2,205 (MNIST), and 2,880 (COIL20) , respectively.
  \item LBP: Like DSIFT, we divide each image into multiple non-overlapping patches and then extract LBP features using 8 sampling points on a circle of radius 1. Thus, we obtain a 59-dimensional LBP feature vector from each patch. By concatenating the descriptors of each image, we obtain a feature vector with the dimension of 7,788 (COIL20) and 2,891 (MNIST).
  \item LPQ: The patch size is set as  $8\times 8$ for COIL20 and MNIST. For all the tested data sets, we set the size of LPQ window as 3, 5, and 7. By concatenating the features of all patches of each image, the dimension of each feature is 12,288 for COIL20 and 6,912 for MNIST.
\end{itemize}
For computational efficiency, we perform PCA to reduce the feature dimension of all data sets to 300, by following the setting in \cite{Hu2014:Dis,Elhamifar2013}

\textbf{Baseline Methods:}
We compare DSSC with 12 state-of-the-art subspace clustering methods, \textit{i.e.} SSC~\cite{Elhamifar2009,Elhamifar2013}, Kernel SSC (KSSC)~\cite{Patel2014:Image}, LRR~\cite{Liu2010,Liu2013}, low rank subspace clustering (LRSC)~\cite{Favaro2011}, Kernel LRR~\cite{Xiao2015:Robust}, least square regression (LSR)~\cite{Lu2012}, smooth representation (SMR)~\cite{Hu2014:Smooth}. LSR has two variants which are denoted by LSR1 and LSR2, respectively. KSSC and KLRR have also two variants which are based on the RBF function (KSSC1 / KLRR1) and the polynomial function (KSSC2 / KLRR2), respectively. Moreover, we have also used the deep autoencoder (DAE) with SSC as a baseline to show the efficacy of our method. More specifically, we adopt the pre-training and fine-tuning strategy~\cite{Hinton2006:Reducing} to train a DAE that consists of five layers with 300, 200, 150, 200, and 300 neurons. In the experiments, we investigate the performance of DAE with two popular nonlinear activation functions, \textit{i.e.} the sigmoid function (DAEg) and the saturating linear transfer function (DAEs). After the DAE converging, we perform SSC on the output of the third layer to obtain the clustering results. For fair comparisons, we use the same $\ell_1$-solver (\textit{i.e.} the Homotopy method~\cite{Yang2010,Osborne2000}) to solve the $\ell_1$-minimization problem in DSSC, SSC, and DAE. Noted that, our method could also be compatible to other neural networks such as convolutional neural networks (CNN). In experiments, we adopt the fully connected network (FCN) instead of CNN because the former has offered a desirable performance in our experiments. Moreover, FCN is with fewer hyper-parameters than CNN, which remarkably reduces the effort to seek optimal value for hyper-parameters.

\begin{table*}[!t]
\caption{Clustering results on the \textbf{COIL20} data set. Results in boldface are significantly better than the others, according to the t-test with a significance level at 0.05.}
\label{tab:1}
\centering
\begin{scriptsize}
\begin{tabular}{l| lllll| lllll }
\toprule
\multicolumn{1}{c|}{Features} & \multicolumn{5}{c|}{DSIFT} & \multicolumn{5}{c}{HOG} \\
\hline
\multicolumn{1}{c|}{Methods} & \multicolumn{1}{c}{ACC} & \multicolumn{1}{c}{NMI} & \multicolumn{1}{c}{ARI} & \multicolumn{1}{c}{Fscore} & \multicolumn{1}{c|}{Para.} & \multicolumn{1}{c}{ACC} & \multicolumn{1}{c}{NMI} & \multicolumn{1}{c}{ARI} & \multicolumn{1}{c}{Fscore} & \multicolumn{1}{c}{Para.}\\
\midrule
DSSC & \textbf{80.82$\pm$2.88} & \textbf{90.52$\pm$0.94} & \textbf{77.63$\pm$2.09} & \textbf{78.88$\pm$1.96} & $2^{-12}$, 20 & \textbf{87.10$\pm$2.82} & \textbf{91.67$\pm$1.07} & \textbf{82.56$\pm$1.26} & \textbf{83.51$\pm$2.12} & $2^{-12}$, 30\\
SSC & 78.96$\pm$3.12 & 89.06$\pm$1.03 & 76.46$\pm$2.31 & 77.59$\pm$2.17 & 0.5, 0.2 & 85.01$\pm$0.85 & 89.99$\pm$0.38 & 81.13$\pm$1.08 & 82.08$\pm$1.02 & 0.5, 0.1\\
KSSC1 & 71.00$\pm$2.13 & 78.72$\pm$0.98 & 63.33$\pm$1.85 & 65.18$\pm$1.75 & $10^{-3}$, 18 & 75.29$\pm$0.97 & 82.75$\pm$0.49 & 66.46$\pm$1.43 & 68.20$\pm$1.33 & $10^{-2}$, 18\\
KSSC2 & 72.01$\pm$2.68 & 83.84$\pm$0.89 & 64.22$\pm$3.47 & 66.22$\pm$3.16 & $10^{-3}$, 18 & 69.53$\pm$1.30 & 81.27$\pm$0.69 & 61.16$\pm$1.83 & 63.32$\pm$1.69 & $10^{-2}$, 18\\
DAEg & 55.83$\pm$2.80 & 70.42$\pm$1.43 & 47.06$\pm$2.74 & 50.00$\pm$2.52 & 0.5, 0.2 & 69.60$\pm$1.00 & 78.52$\pm$0.47 & 59.38$\pm$0.79 & 61.63$\pm$0.74 & 0.5, 0.1\\
DAEs & 55.81$\pm$2.60 & 70.71$\pm$1.68 & 48.49$\pm$3.31 & 51.46$\pm$3.05 & 0.5, 0.2 & 64.75$\pm$1.31 & 77.48$\pm$0.60 & 56.81$\pm$1.12 & 59.13$\pm$1.06 & 0.5, 0.1\\
LRR & 71.03$\pm$1.47 & 80.52$\pm$1.05 & 63.83$\pm$2.09 & 65.70$\pm$1.97 & 5e-2  & 76.89$\pm$1.46 & 86.52$\pm$0.78 & 70.79$\pm$1.73 & 72.39$\pm$1.62 & 5e-3\\
KLRR1 & 70.46$\pm$1.55 & 79.61$\pm$1.01 & 61.25$\pm$1.94 & 63.35$\pm$1.81 & 500  & 76.74$\pm$0.27 & 82.00$\pm$0.14 & 69.43$\pm$0.48 & 70.96$\pm$0.45 & 10\\
KLRR2 & 70.85$\pm$1.37 & 80.09$\pm$1.15 & 62.75$\pm$1.54 & 64.73$\pm$1.46 & 100 & 72.33$\pm$2.65 & 80.98$\pm$1.21 & 63.11$\pm$2.88 & 65.07$\pm$2.68 & 5\\
LRSC & 71.82$\pm$0.28 & 77.65$\pm$0.23 & 62.72$\pm$0.52 & 64.62$\pm$0.49 & 0.08 & 57.11$\pm$1.24 & 69.91$\pm$0.73 & 46.27$\pm$1.57 & 49.20$\pm$1.48 & 0.01\\
LSR1 & 63.93$\pm$2.15 & 73.18$\pm$1.12 & 53.29$\pm$2.26 & 55.75$\pm$2.14 & 0.6 & 54.81$\pm$1.80 & 64.44$\pm$0.94 & 42.28$\pm$1.55 & 45.35$\pm$1.44 & 0.5\\
LSR2 & 68.11$\pm$1.14 & 75.33$\pm$0.62 & 56.29$\pm$1.56 & 58.61$\pm$1.41 & 0.9  & 53.81$\pm$1.51 & 63.00$\pm$1.22 & 42.07$\pm$1.5 & 45.19$\pm$1.42 & 0.3\\
SMR & 76.97$\pm$0.96 & 85.30$\pm$0.58 & 71.56$\pm$1.02 & 73.02$\pm$0.96 & $2^{-16}$,$10^{-3}$ & 80.15$\pm$0.87 & 85.93$\pm$0.6 & 73.51$\pm$1.06 & 74.87$\pm$1.01 & $2^{-16}$,$10^{-3}$\\
\bottomrule
\end{tabular}
\end{scriptsize}
\end{table*}

\begin{table*}[!t]
\caption{Clustering results on the \textbf{COIL20} data set. Results in boldface are significantly better than the others, according to the t-test with a significance level at 0.05.}
\label{tab:2}
\centering
\begin{scriptsize}
\begin{tabular}{l| lllll| lllll }
\toprule
\multicolumn{1}{c|}{Features} & \multicolumn{5}{c|}{LBP} & \multicolumn{5}{c}{LPQ} \\
\hline
\multicolumn{1}{c|}{Methods} & \multicolumn{1}{c}{ACC} & \multicolumn{1}{c}{NMI} & \multicolumn{1}{c}{ARI} & \multicolumn{1}{c}{Fscore} & \multicolumn{1}{c|}{Para.} & \multicolumn{1}{c}{ACC} & \multicolumn{1}{c}{NMI} & \multicolumn{1}{c}{ARI} & \multicolumn{1}{c}{Fscore} & \multicolumn{1}{c}{Para.}\\
\midrule
DSSC & \textbf{72.89$\pm$1.41} & \textbf{84.32$\pm$0.79} & \textbf{67.31$\pm$1.96} & \textbf{69.01$\pm$1.85} & $2^{-13}$, 40 & \textbf{78.12$\pm$2.09} & \textbf{85.38$\pm$0.77} & \textbf{71.35$\pm$1.34} & \textbf{72.87$\pm$1.25} & $2^{-12}$, 60\\
SSC & 70.17$\pm$0.65 & 82.66$\pm$0.19 & 64.19$\pm$0.60 & 66.07$\pm$0.58 & $10^{-3}$, $10^{-2}$ & 74.60$\pm$0.81 & 84.21$\pm$0.49 & 67.69$\pm$0.83 & 69.35$\pm$0.79 & $10^{-3}$, 0.1\\
KSSC1 & 69.33$\pm$1.97 & 80.65$\pm$0.86 & 61.15$\pm$1.91 & 63.18$\pm$1.79 & 1, 16 & 68.49$\pm$2.38 & 79.28$\pm$1.27 & 59.06$\pm$2.37 & 61.23$\pm$2.21 & 0.1, 12\\
KSSC2 & 70.42$\pm$1.13 & 83.67$\pm$0.69 & 65.28$\pm$1.23 & 68.03$\pm$1.16 & 1, 16 & 69.24$\pm$2.33 & 79.52$\pm$0.93 & 61.07$\pm$1.72 & 63.17$\pm$1.62 & 0.1, 12\\
DAEg & 40.96$\pm$2.18 & 53.54$\pm$0.89 & 26.27$\pm$1.33 & 30.57$\pm$1.22 & $10^{-3}$, $10^{-2}$  & 62.19$\pm$0.90 & 72.04$\pm$0.54 & 51.51$\pm$0.75 & 54.15$\pm$0.72 & $10^{-3}$,  0.1\\
DAEs & 40.68$\pm$1.13 & 52.12$\pm$0.92 & 23.67$\pm$1.30 & 28.26$\pm$1.10 & $10^{-3}$, $10^{-2}$  & 59.64$\pm$2.46 & 67.44$\pm$1.06 & 44.90$\pm$1.81 & 47.98$\pm$1.65 & $10^{-3}$,  0.1\\
LRR & 71.60$\pm$4.02 & 84.45$\pm$1.78 & 65.47$\pm$5.68 & 66.29$\pm$5.21 & 0.5  & 69.00$\pm$1.09 & 80.31$\pm$0.88 & 60.12$\pm$1.51 & 62.29$\pm$1.41 & 0.1\\
KLRR1 & 65.83$\pm$0.31 & 77.34$\pm$0.30 & 56.41$\pm$0.50 & 58.60$\pm$0.47 & 30  & 69.43$\pm$1.46 & 77.34$\pm$0.53 & 57.01$\pm$1.02 & 59.24$\pm$0.96 & 500\\
KLRR2 & 70.10$\pm$1.27 & 79.58$\pm$0.13 & 62.91$\pm$0.51 & 64.82$\pm$0.47 & 1000  & 65.33$\pm$2.48 & 76.41$\pm$1.13 & 54.22$\pm$2.11 & 56.69$\pm$1.94 & 100\\ 
LRSC & 62.96$\pm$0.61 & 73.38$\pm$0.79 & 53.31$\pm$1.06 & 55.67$\pm$1.01 & 0.04 & 66.38$\pm$0.50 & 78.73$\pm$0.58 & 58.81$\pm$0.97 & 60.99$\pm$0.91 & 0.08\\
LSR1 & 70.24$\pm$2.90 & 82.40$\pm$1.41 & 64.54$\pm$2.85 & 67.33$\pm$2.69 & 1  & 66.97$\pm$1.68 & 74.42$\pm$0.62 & 55.48$\pm$1.52 & 57.74$\pm$1.43 & 0.2\\
LSR2 & 70.54$\pm$3.26 & 81.63$\pm$1.16 & 63.71$\pm$2.58 & 66.59$\pm$2.41 & 0.6  & 65.25$\pm$1.55 & 73.81$\pm$1.29 & 54.34$\pm$1.65 & 56.66$\pm$1.56 & 0.3\\
SMR & 71.93$\pm$1.35 & 81.17$\pm$0.39 & 63.54$\pm$1.41 & 66.39$\pm$1.32 & $2^{-16}$, $10^{-3}$ & 70.56$\pm$0.57 & 80.68$\pm$0.41 & 61.68$\pm$0.49 & 63.68$\pm$0.45 & $2^{-16}$, $10^{-3}$\\
%\hline
%\multicolumn{1}{c|}{\multirow{3}{*}{Overview}} & \multicolumn{3}{l|}{DSSC $\succ$ SSC $\succ$ DAEg $\succ$ KSSC1 $\succ$ LRR $\succ$  } & \multicolumn{3}{l}{DSSC $\succ$ DAEs $\succ$ SSC $\succ$ DAEg $\succ$ KSSC1 $\succ$  }\\
%& \multicolumn{3}{r|}{SMR $\succ$ DAEs $\succ$ KSSC2 $\succ$ LSR2 $\succ$ LSR1 $\succ$ } & \multicolumn{3}{r}{LRR $\succ$ SMR $\succ$ LSR2 $\succ$ LRSC $\succ$ LSR1 $\succ$}\\
%& \multicolumn{3}{r|}{KLRR1 $\succ$ LRSC $\succ$ KLRR2} & \multicolumn{3}{r|}{KLRR1 $\succ$ KSSC2 $\succ$ KLRR2}\\
\bottomrule
\end{tabular}
\end{scriptsize}
\end{table*}

\begin{table*}[!t]
\caption{Clustering results on the \textbf{MNIST} data set. Results in boldface are significantly better than the others, according to the t-test with a significance level at 0.05.}
\label{tab:3}
\centering
\begin{scriptsize}
\begin{tabular}{l| lllll| lllll }
\toprule
\multicolumn{1}{c|}{Features} & \multicolumn{5}{c|}{DSIFT} & \multicolumn{5}{c}{HOG} \\
\hline
\multicolumn{1}{c|}{Methods} & \multicolumn{1}{c}{ACC} & \multicolumn{1}{c}{NMI} & \multicolumn{1}{c}{ARI} & \multicolumn{1}{c}{Fscore} & \multicolumn{1}{c|}{Para.} & \multicolumn{1}{c}{ACC} & \multicolumn{1}{c}{NMI} & \multicolumn{1}{c}{ARI} & \multicolumn{1}{c}{Fscore} & \multicolumn{1}{c}{Para.}\\
\midrule
DSSC & \textbf{72.65$\pm$0.00} & \textbf{70.42$\pm$0.00} & \textbf{61.80$\pm$0.00} & \textbf{65.79$\pm$0.00} & $2^{-13}$, 20 & \textbf{78.10$\pm$0.00} & \textbf{77.51$\pm$0.00} & \textbf{68.72$\pm$0.00} & \textbf{72.03$\pm$0.00} & $2^{-17}$, 30\\
SSC & 62.45$\pm$0.00 & 65.75$\pm$0.00 & 53.75$\pm$0.00 & 58.81$\pm$0.00 & 1, $10^{-2}$ & 77.35$\pm$0.00 & 75.70$\pm$0.00 & 66.90$\pm$0.00 & 70.23$\pm$0.00 & 10, $10^{-2}$\\
KSSC1 & 50.90$\pm$0.00 & 49.75$\pm$0.00 & 35.28$\pm$0.00 & 41.80$\pm$0.00 & $10^{-3}$, 10 & 66.90$\pm$0.00 & 70.20$\pm$0.00 & 56.79$\pm$0.00 & 61.35$\pm$0.00 & $10^{-2}$, 12\\
KSSC2 & 60.80$\pm$0.00 & 63.96$\pm$0.00 & 50.81$\pm$0.00 & 56.26$\pm$0.00 & $10^{-3}$, 10 & 68.00$\pm$0.00 & 70.74$\pm$0.00 & 57.69$\pm$0.00 & 62.12$\pm$0.00 & $10^{-2}$, 12\\
DAEg & 52.55$\pm$0.00 & 58.36$\pm$0.00 & 40.98$\pm$0.00 & 47.48$\pm$0.00 & 1, $10^{-2}$ & 23.36$\pm$0.66 & 11.56$\pm$0.78 & 5.83$\pm$0.28 & 15.89$\pm$0.32 & 10, $10^{-2}$\\
DAEs & 42.28$\pm$1.19 & 48.70$\pm$0.56 & 31.70$\pm$0.31 & 39.85$\pm$0.22 & 1, $10^{-2}$ & 23.07$\pm$1.14 & 10.48$\pm$0.52 & 4.91$\pm$0.31 & 15.02$\pm$0.29 & 10, $10^{-2}$\\
LRR & 63.20$\pm$0.00 & 68.34$\pm$0.00 & 54.11$\pm$0.00 & 59.48$\pm$0.00 & 0.05  & 73.30$\pm$0.00 & 74.49$\pm$0.00 & 63.20$\pm$0.00 & 67.12$\pm$0.00 & 0.01\\
KLRR1 & 57.05$\pm$0.00 & 57.97$\pm$0.00 & 44.63$\pm$0.00 & 50.70$\pm$0.00 & 3000 & 72.15$\pm$0.00 & 70.95$\pm$0.00 & 61.38$\pm$0.00 & 65.42$\pm$0.00 & 30\\
KLRR2 & 22.63$\pm$0.89 & 12.55$\pm$1.37 & 9.02$\pm$0.46 & 22.98$\pm$0.14 & 1000  & 73.55$\pm$0.00 & 73.30$\pm$0.00 & 63.69$\pm$0.00 & 67.50$\pm$0.00 & 3\\
LRSC & 59.30$\pm$0.00 & 58.84$\pm$0.00 & 46.90$\pm$0.00 & 52.37$\pm$0.00 & 0.1 & 61.20$\pm$0.00 & 59.65$\pm$0.00 & 47.05$\pm$0.00 & 52.59$\pm$0.00 & 0.01\\
LSR1 & 63.50$\pm$0.00 & 60.39$\pm$0.00 & 49.02$\pm$0.00 & 54.29$\pm$0.00 & 0.1  & 58.42$\pm$0.09 & 56.41$\pm$0.07 & 44.85$\pm$0.12 & 50.79$\pm$0.11 & 0.2\\
LSR2 & 63.55$\pm$0.00 & 60.53$\pm$0.00 & 49.14$\pm$0.00 & 54.39$\pm$0.00 & 0.4  & 60.40$\pm$0.02 & 57.78$\pm$0.00 & 46.45$\pm$0.00 & 51.98$\pm$0.00 & 0.1\\
SMR & 69.15$\pm$0.00 & 68.90$\pm$0.00 & 59.17$\pm$0.00 & 63.40$\pm$0.00 & $2^{-16}$, $10^{-3}$ & 77.22$\pm$0.05 & 77.25$\pm$0.00 & 66.85$\pm$0.00 & 71.11$\pm$0.00 & $2^{-16}$, $10^{-3}$\\
\bottomrule
\end{tabular}
\end{scriptsize}
\end{table*}

\begin{table*}[!t]
\caption{Clustering results on the \textbf{MNIST} data set. Results in boldface are significantly better than the others, according to the t-test with a significance level at 0.05.}
\label{tab:4}
\centering
\begin{scriptsize}
\begin{tabular}{l| lllll| lllll }
\toprule
\multicolumn{1}{c|}{Features} & \multicolumn{5}{c|}{LBP} & \multicolumn{5}{c}{LPQ} \\
\hline
\multicolumn{1}{c|}{Methods} & \multicolumn{1}{c}{ACC} & \multicolumn{1}{c}{NMI} & \multicolumn{1}{c}{ARI} & \multicolumn{1}{c}{Fscore} & \multicolumn{1}{c|}{Para.} & \multicolumn{1}{c}{ACC} & \multicolumn{1}{c}{NMI} & \multicolumn{1}{c}{ARI} & \multicolumn{1}{c}{Fscore} & \multicolumn{1}{c}{Para.}\\
\midrule
DSSC & \textbf{61.70$\pm$0.00} & \textbf{54.23$\pm$0.00} & \textbf{44.54$\pm$0.00} & \textbf{50.25$\pm$0.00} & $2^{-15}$, 40 & \textbf{65.04$\pm$0.02} & \textbf{54.85$\pm$0.01} & \textbf{46.38$\pm$0.00} & \textbf{52.04$\pm$0.00} & $2^{-16}$, 70\\
SSC & 59.75$\pm$0.00 & 53.83$\pm$0.00 & 43.52$\pm$0.00 & 49.02$\pm$0.00 & 0.1, 0.01 & 62.35$\pm$0.00 & 53.86$\pm$0.00 & 44.67$\pm$0.00 & 50.42$\pm$0.00 & 1, 0.01\\
KSSC1 & 58.50$\pm$0.00 & 53.96$\pm$0.00 & 41.58$\pm$0.00 & 47.67$\pm$0.00 & 0.1, 10 & 44.00$\pm$0.00 & 34.97$\pm$0.00 & 23.49$\pm$0.00 & 31.26$\pm$0.00 & $10^{-3}$, 16\\
KSSC2 & 57.70$\pm$0.00 & 54.27$\pm$0.00 & 41.95$\pm$0.00 & 47.96$\pm$0.00 & 0.1, 10 & 54.99$\pm$0.03 & 51.07$\pm$0.01 & 36.62$\pm$0.01 & 43.21$\pm$0.01 & $10^{-3}$, 16\\
DAEg & 36.20$\pm$0.00 & 27.81$\pm$0.00 & 16.68$\pm$0.00 & 25.42$\pm$0.00 & 0.1, 0.01 & 30.61$\pm$0.51 & 22.05$\pm$0.32 & 12.19$\pm$0.05 & 22.07$\pm$0.05 & 10, 0.01\\
DAEs & 32.20$\pm$0.00 & 24.85$\pm$0.00 & 14.75$\pm$0.00 & 23.39$\pm$0.00 & 0.1, 0.01 & 34.10$\pm$0.00 & 22.71$\pm$0.02 & 12.41$\pm$0.01 & 22.43$\pm$0.01 & 10, 0.01\\
LRR & 55.70$\pm$0.00 & 45.70$\pm$0.00 & 37.77$\pm$0.00 & 44.58$\pm$0.00 & 0.5  & 52.15$\pm$0.00 & 50.63$\pm$0.00 & 37.86$\pm$0.00 & 44.63$\pm$0.00 & 0.5\\
KLRR1 & 54.12$\pm$0.06 & 50.98$\pm$0.01 & 37.84$\pm$0.07 & 44.25$\pm$0.06 & 1000  & 55.60$\pm$0.00 & 51.66$\pm$0.00 & 38.42$\pm$0.00 & 44.77$\pm$0.00 & 3000\\
KLRR2 & 53.75$\pm$0.00 & 50.70$\pm$0.00 & 37.05$\pm$0.00 & 43.55$\pm$0.00 & 300  & 56.75$\pm$0.00 & 51.69$\pm$0.00 & 38.87$\pm$0.00 & 45.16$\pm$0.00 & 1000\\
LRSC & 42.45$\pm$0.00 & 35.60$\pm$0.00 & 23.42$\pm$0.00 & 31.43$\pm$0.00 & 0.03 & 53.20$\pm$0.00 & 42.49$\pm$0.00 & 32.03$\pm$0.00 & 39.04$\pm$0.00 & 0.05\\
LSR1 & 53.12$\pm$0.05 & 45.81$\pm$0.05 & 35.27$\pm$0.01 & 41.87$\pm$0.01 & 0.1  & 52.60$\pm$0.00 & 46.93$\pm$0.00 & 34.93$\pm$0.00 & 41.71$\pm$0.00 & 1\\
LSR2 & 52.93$\pm$0.04 & 45.65$\pm$0.03 & 34.98$\pm$0.05 & 41.61$\pm$0.05 & 0.1  & 53.25$\pm$0.05 & 47.57$\pm$0.11 & 35.54$\pm$0.06 & 42.28$\pm$0.06 & 1\\
SMR & 49.90$\pm$0.00 & 44.29$\pm$0.00 & 32.16$\pm$0.00 & 39.18$\pm$0.00 & $2^{-14}$, $10^{-3}$ & 48.90$\pm$0.00 & 43.43$\pm$0.00 & 30.18$\pm$0.00 & 37.62$\pm$0.00 & $2^{-14}$, $10^{-3}$\\
\bottomrule
\end{tabular}
\end{scriptsize}
\end{table*}

\textbf{Experimental Settings:} In our experiments, we adopt cross-validation for selecting the optimal parameters for all the tested methods~\cite{nntricks:2012}\footnote{The following parameters are tuned with the cross validation technique: DSSC ($\mu$, $\tau$), SSC ($\gamma$, $\delta$), KSSC ($\gamma$, $\delta$), DAE ($\gamma$, $\delta$), LRR ($\lambda$), KLRR ($\lambda$), LRSC ($\lambda$), LSR ($\lambda$), and SMR ($\alpha$, $k$).}. More specifically, we equally split each data set into two partitions and tune parameters using one partition. With the tuned parameters, we repeat each algorithm 10 times on the other partition and report the achieved mean and standard deviation of the used clustering performance metrics. In all the experiments, we train a DSSC consisting of three layers, with 300, 200, and 150 neurons respectively. Moreover, we set $\lambda =10^{-3}/n$ and the convergence threshold as $10^{-3}$ for DSSC and adopt early stopping technique (\textit{w.r.t.} the parameter $\tau$) to avoid overfitting by following~\cite{nntricks:2012}, where $n$ is the data size. Once the network converges, we experimentally found that removing the nonlinear functions could be helpful for following clustering step in inference phrase. Note that, we directly use the tuned parameters $\gamma$ (sparsity) and $\delta$ (tolerance) of SSC for DSSC. If these two parameters are tuned specifically, the performance of DSSC could be further improved. 

\textbf{Evaluation Criteria:} Like~\cite{Zhang:2015ku}, we adopt four popular metrics to evaluate the clustering performance of our algorithm, \textit{i.e.} accuracy (\textit{ACC}) or called \textit{purity}, normalized mutual information (\textit{NMI}), adjusted rand index (\textit{ARI}), and Fscore. Higher value of these metrics indicates better performance. 

%\begin{figure}[!t]
%\begin{center}
%\subfigure [Activation functions]{\label{fig2a}\includegraphics[width=0.236\textwidth]{fig2a.pdf}}
%\subfigure [The derivate of activation functions]{\label{fig2b}\includegraphics[width=0.236\textwidth]{fig2b.pdf}}
%\end{center}
%\caption{\label{fig2} A illustration example to show the used activation function and the corresponding derivate.}
%\end{figure}

%\def \figwidth {0.49}
%\begin{figure*}[t]
%\begin{center}
%\subfigure [DSIFT]{\label{fig2a}\includegraphics[width=\figwidth\textwidth]{fig2a.pdf}}\hspace{1mm}
%\subfigure [SDISFT]{\label{fig2b}\includegraphics[width=\figwidth\textwidth]{fig2b.pdf}}\\
%\subfigure [HOG]{\label{fig2c}\includegraphics[width=\figwidth\textwidth]{fig2c.pdf}}\hspace{1mm}
%\subfigure [LBP]{\label{fig2d}\includegraphics[width=\figwidth\textwidth]{fig2d.pdf}}
%\end{center}
%\caption{\label{fig2} Performance comparisons with the state-of-the-art subspace clustering methods on the \textbf{MNIST} data sets in terms of \textbf{ACC}. 
%}
%\end{figure*}

\subsection{Comparison with state-of-the-art methods}

In this section, we compare DSSC with 12 recently-proposed subspace clustering methods on the COIL20 and the MNIST data sets, where each data set is with four different features.

\textbf{On COIL20:} We first investigate the performance of DSSC using the COIL20 data set. Tables~\ref{tab:1}--\ref{tab:2} report the results from which we can see that:
\begin{itemize}
 \item DSSC consistently outperforms  other tested methods in terms of all of the used performance metrics. Regarding the four types of features, DSSC achieves at least 1.86\%, 2.09\%, 0.96\% and 3.52\% relative improvement over the \textit{ACC} of the best baseline, respectively. 
  \item SSC usually outperforms DAEs and DAEg, whereas our DSSC method consistently outperforms SSC in all the settings. This shows that it is hard to achieve a desirable performance by simply introducing deep learning into subspace clustering since unsupervised deep learning is an open challenging issue~\cite{Bengio2013:Rep}. 
\end{itemize}

\textbf{On MNIST:} We also investigate the performance of DSSC by using the MNIST data set. 

Tables~\ref{tab:3}--\ref{tab:4} show the result, from which we obverse that the \textit{ACC} of DSSC with the DSIFT feature is 72.65\% which improves SSC by 10.20\% and the best baseline algorithm by 3.50\%. With respect to the other three features, the improvement of DSSC comparing with all the baseline approaches is also significant, which is 1.82\%, 1.02\%, and 1.71\% in terms of \textit{ARI}. It should be pointed out that, all the tested methods perform very stable on this data set, whose standard deviations on these four performance metrics are close to 0. 

\begin{table*}[!t]
\caption{Deep vs. Shallow Models on the \textbf{AR} data set. Results in boldface are significantly better than the others, according to the t-test with a significance level at 0.05.}
\label{tab:5}
\centering
\begin{scriptsize}
\begin{tabular}{l| lllll| lllll }
\toprule
\multicolumn{1}{c|}{Features} & \multicolumn{5}{c|}{DSIFT} & \multicolumn{5}{c}{HOG} \\
\hline
\multicolumn{1}{c|}{Methods} & \multicolumn{1}{c}{ACC} & \multicolumn{1}{c}{NMI} & \multicolumn{1}{c}{ARI} & \multicolumn{1}{c}{Fscore} & \multicolumn{1}{c|}{Para.} & \multicolumn{1}{c}{ACC} & \multicolumn{1}{c}{NMI} & \multicolumn{1}{c}{ARI} & \multicolumn{1}{c}{Fscore} & \multicolumn{1}{c}{Para.}\\
\midrule
DSSC(M=2) & \textbf{85.38$\pm$1.08} & \textbf{95.17$\pm$0.17 }& \textbf{82.15$\pm$0.63} & \textbf{82.35$\pm$0.62} & $2^{-11}$, 50 & \textbf{85.05$\pm$1.53} & \textbf{94.36$\pm$0.43} & \textbf{78.98$\pm$1.58} & \textbf{79.21$\pm$1.56} & $2^{-12}$, 30\\
DSSC (M=1) & 83.81$\pm$1.72 & 94.57$\pm$0.45 & 81.23$\pm$1.94 & 81.42$\pm$1.92 & $2^{-11}$, 30 & 81.90$\pm$0.96 & 91.93$\pm$0.35 & 71.87$\pm$1.97 & 72.17$\pm$1.95 & $2^{-12}$, 20\\
SSC & 74.83$\pm$1.27 & 89.91$\pm$0.38 & 66.43$\pm$1.44 & 66.81$\pm$1.42 & $10^{-2}$, $10^{-3}$ & 81.65$\pm$1.18 & 92.48$\pm$0.41 & 74.23$\pm$1.76 & 74.52$\pm$1.74 & 0.5, $10^{-3}$\\
KSSC1 & 70.27$\pm$1.66 & 87.29$\pm$0.53 & 58.61$\pm$1.78 & 59.08$\pm$1.76 & 1, 18 & 83.12$\pm$0.90 & 93.07$\pm$0.34 & 75.68$\pm$1.37 & 75.94$\pm$1.36 & $10^{-2}$, 20\\
KSSC2 & 78.28$\pm$1.78 & 91.55$\pm$0.39 & 71.13$\pm$1.44 & 71.44$\pm$1.43 & 1, 18 & 83.22$\pm$1.34 & 92.71$\pm$0.32 & 74.56$\pm$1.06 & 74.84$\pm$1.05 & $10^{-2}$, 20\\
DAEg & 74.37$\pm$1.20 & 89.53$\pm$0.43 & 65.42$\pm$1.56 & 65.81$\pm$1.54 & $10^{-2}$, $10^{-3}$ & 74.67$\pm$1.25 & 89.07$\pm$0.49 & 63.77$\pm$1.52 & 64.17$\pm$1.50 & 0.5, $10^{-3}$\\
DAEs & 72.65$\pm$0.91 & 88.54$\pm$0.52 & 62.23$\pm$1.81 & 62.67$\pm$1.78 & $10^{-2}$, $10^{-3}$ & 73.32$\pm$1.31 & 88.17$\pm$0.43 & 61.12$\pm$1.42 & 61.56$\pm$1.40 & 0.5, $10^{-3}$\\
LRR & 82.67$\pm$1.00 & 93.48$\pm$0.33 & 77.33$\pm$1.37 & 77.60$\pm$1.35 & 0.1  & 83.00$\pm$1.36 & 93.27$\pm$0.46 & 77.34$\pm$3.21 & 77.61$\pm$3.16 & 0.01\\
KLRR1 & 79.92$\pm$1.52 & 91.56$\pm$0.51 & 71.08$\pm$2.13 & 71.42$\pm$2.10 & 300 & 83.92$\pm$1.26 & 93.00$\pm$0.45 & 77.49$\pm$1.49 & 77.73$\pm$1.47 & 100\\
KLRR2 & 23.08$\pm$0.36 & 52.01$\pm$0.62 & 5.31$\pm$0.24 & 6.73$\pm$0.24 & 100  & 76.07$\pm$1.69 & 88.78$\pm$0.74 & 63.93$\pm$2.67 & 64.34$\pm$2.63 & 5\\
LRSC & 83.55$\pm$1.20 & 92.84$\pm$0.37 & 78.33$\pm$1.39 & 78.57$\pm$1.38 & 0.06 & 83.42$\pm$1.43 & 92.67$\pm$0.48 & 73.86$\pm$1.73 & 74.15$\pm$1.71 & 0.02\\
LSR1 & 82.43$\pm$1.31 & 92.69$\pm$0.49 & 74.94$\pm$1.87 & 75.22$\pm$1.85 & 0.3  & 83.32$\pm$1.70 & 92.45$\pm$0.49 & 73.11$\pm$2.24 & 73.40$\pm$2.21 & 0.8\\
LSR2 & 82.45$\pm$1.58 & 92.64$\pm$0.42 & 74.49$\pm$1.80 & 74.77$\pm$1.78 & 0.7  & 83.65$\pm$1.07 & 92.45$\pm$0.45 & 73.24$\pm$1.77 & 73.54$\pm$1.75 & 1\\
SMR & 71.07$\pm$1.91 & 87.01$\pm$0.52 & 60.82$\pm$2.22 & 61.26$\pm$2.19 & $2^{-16}$, $10^{-2}$ & 81.38$\pm$0.73 & 91.75$\pm$0.27 & 72.51$\pm$0.85 & 72.81$\pm$0.84 & $2^{-15}$, $10^{-2}$\\
\bottomrule
\end{tabular}
\end{scriptsize}
\end{table*}

\subsection{Deep Model vs. Shallow Models}

In this section, we investigate the influence of the depth of DSSC using the AR data set with  DSIFT and HOG features. More specifically, we report the performance of DSSC with two hidden layers ($M=2$) and one hidden layer ($M=1$), respectively. In the case of $M=1$, the number of hidden neurons is also set as 150. Note that, KSSC1 and KSSC2 can be regarded as two shallow models of SSC with one nonlinear hidden layer.

Table~\ref{tab:5} shows the clustering results of the methods, as well as the tuned parameters. We observe that our DSSC ($M=2$) consistently outperform the shallow models in terms of all of these evaluation metrics. The results also verify our claim and motivation, \textit{i.e.} our deep model DSSC significantly benefit from deep learning.

%\def \ourwidth {0.46}
%\begin{figure*}[!t]
%\begin{center}
%\subfigure [ACC]{\label{fig2a}\includegraphics[width=\ourwidth\textwidth]{fig2a.pdf}}
%\subfigure [NMI]{\label{fig2b}\includegraphics[width=\ourwidth\textwidth]{fig2b.pdf}}\\
%\subfigure [ARI]{\label{fig2c}\includegraphics[width=\ourwidth\textwidth]{fig2c.pdf}}
%\subfigure [Fscore]{\label{fig2d}\includegraphics[width=\ourwidth\textwidth]{fig2d.pdf}}
%\end{center}
%\caption{\label{fig2} The influence of different activation functions of DSSC on the BF0502 database. The first four results in each figure are achieved by DSSC with different activation functions.
%}
%\end{figure*}

\begin{table}[!t]
\caption{The influence of different activation functions of DSSC on the BF0502 database. DSSC (t), DSSC (s), DSSC (n), and DSSC (r) denote DSSC with the tanh, sigmoid, nssigmoid, and relu function, respectively.}
\label{tab:3}
\centering
\begin{small}
\begin{tabular}{l| lllll }
\toprule
\multicolumn{1}{c|}{Methods} & \multicolumn{1}{c}{Accuracy} & \multicolumn{1}{c}{NMI}& \multicolumn{1}{c}{ARI}& \multicolumn{1}{c}{Fscore}& \multicolumn{1}{c}{Para.}\\
\midrule
DSSC (t) & 79.50 & 71.02 & 65.11 & 71.09 & $2^{-13}$,90\\
DSSC (s) & \textbf{82.67} & \textbf{79.01} & \textbf{71.69} & 66.55 & $2^{-17}$,60\\
DSSC (n) & 75.08 & 67.72 & 59.17 & \textbf{72.11} & $2^{-17}$,10\\
DSSC (r) & 80.08 & 75.60 & 65.67 & \textbf{72.11} & $2^{-17}$,10\\
\hline
%DSSC w.o.& 75.83 & 71.96 & 68.56 & 69.29 & 1,0.2\\
SSC & 79.50 & 74.83 & 62.37 & 69.15 & 1,0.2\\
KSSC1 & 74.50 & 71.99 & 61.95 & 68.85 & 0.1,12\\
KSSC2 & 77.83 & 69.89 & 70.65 & 70.55 & 0.1,12\\
DAEg & 55.50 & 38.16 & 30.69 & 43.15 & -\\
DAEs & 21.67 & 6.07 & 0.85 & 28.65 &-\\
LRR & 78.17 & 74.89 & 70.57 & 70.58 & 0.01\\
KLRR1 & 75.33 & 66.60 & 56.83 & 64.07 & 3\\
KLRR2 & 75.00 & 69.32 & 68.35 & 74.16 & 3\\
LRSC & 69.17 & 60.60 & 53.28 & 61.71 & 0.01\\
LSR1 & 67.50 & 57.53 & 51.36 & 60.19 & 1.00\\
LSR2 & 77.00 & 59.91 & 56.27 & 63.60 & 0.50\\
SMR & 76.00 & 74.69 & 58.09 & 71.87 & $2^{-16}$,1e-02\\
\bottomrule
\end{tabular}
\end{small}
\end{table}

\subsection{Influence of Different Activation Functions}

In this section, we investigate the influence of different nonlinear activation functions in DSSC. The investigated functions are \textit{sigmoid}, \textit{non-saturating sigmoid} (\textit{nssigmoid}), and the rectified linear unit (relu)~\cite{Nair2010:ReLU}. We carry out experiment on the BF0502 data set which contains facial images detected from the TV series ``Buffy the Vampire Slayer''. Following~\cite{Xiao2015:Robust}, a subset of BF0502 is used, which includes 17,337 faces in 229 tracks from 6 main casts. Each facial image is represented as a 1,937-dimensional vectors extracted from 13 facial landmark points (\textit{e.g.}, the left and right corners of each eye). In our experiments, we use the first 200 samples from each category, thus resulting in 1,200 images in total.
 
From Table~\ref{tab:3}, we can observe that DSSC with different activation functions outperforms SSC by a considerable performance margin. With the \textit{sigmoid} function, DSSC is about 3.17\%, 4.18\%, 9.32\%, and 2.96\% higher than SSC in terms of \textit{Accuracy}, \textit{NMI}, \textit{ARI}, and \textit{Fscore}, respectively. It is worth noting that although \textit{tanh} is not the best activation function, it is more stable than the other three activation functions in our experiments. Thus, we use the \textit{tanh} function as the activation function for comparisons as shown in the above sections.

\subsection{Convergence Analysis and Time Cost}

\begin{figure}[!t]
\begin{center}
\includegraphics[width=0.46\textwidth]{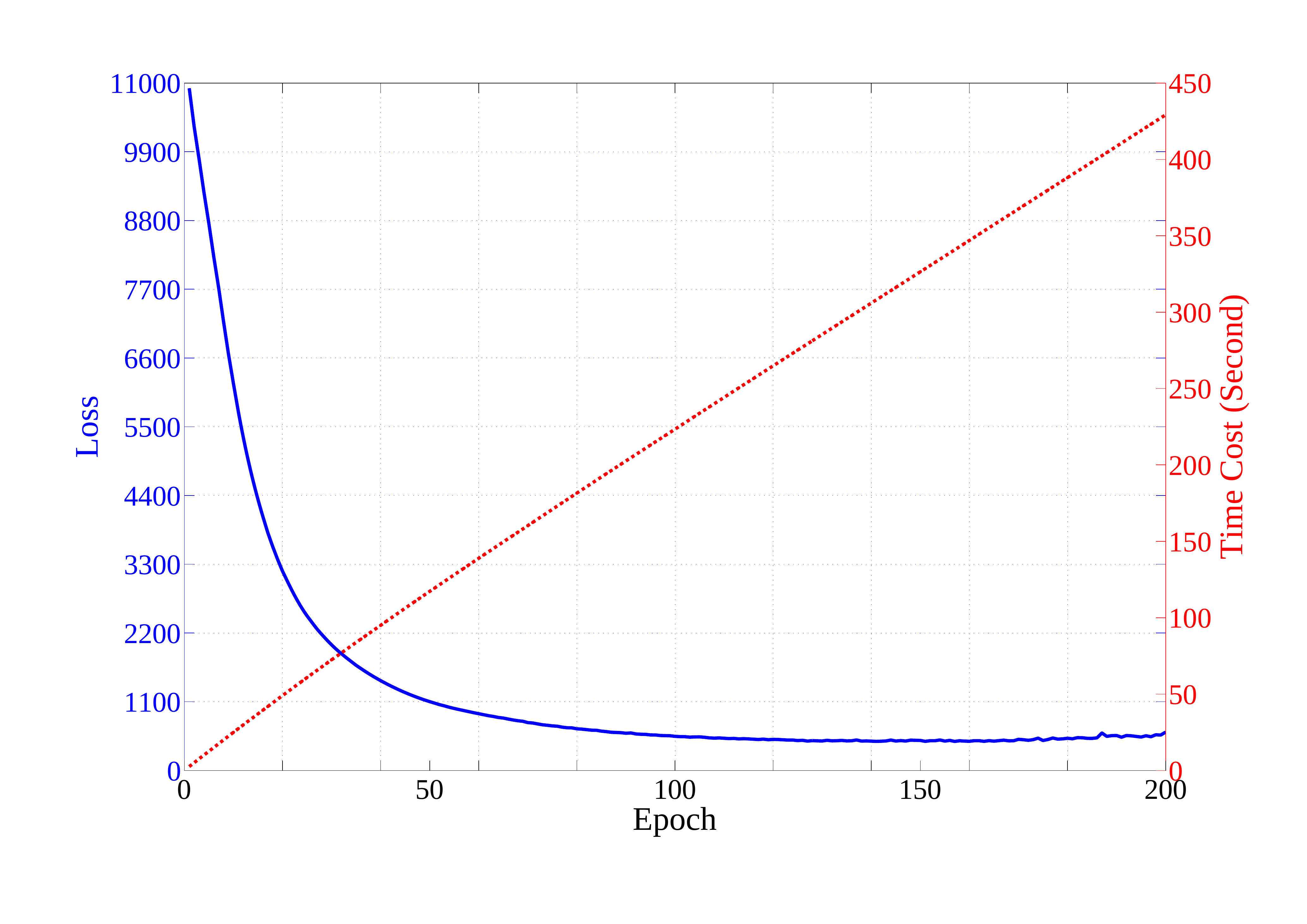}
\end{center}
\caption{\label{fig3} Convergence curve and time cost of DSSC. The left y-axis indicates the loss at each epoch and the right one is the total time cost taken by our method.}
\end{figure}

In this section, we examine the convergence speed and time cost of our DSSC on the BF0502 data set. From Figure~\ref{fig3}, we can see that the loss of DSSC generally keeps unchanged after 90--100 epochs. For each epoch, DSSC takes about 2.2 seconds to obtain results on a macbook with a 2.6GHz Intel Core i5 CPU and 8GB memory. Like other deep learning based methods, the computational cost of DSSC can be remarkably reduced by GPU.

\section{Conclusion}
In this paper, we proposed a new deep learning based framework for simultaneous data representation learning and subspace clustering.  Based on such deep subspace clustering framework, we further devised a new method,  \textit{i.e.} DSSC. Experimental results on the facial, object, and handwritten digit image databases data sets show the efficacy of DSSC in terms of four performance evaluation metrics. In the future, we plan to investigate the performance of our proposed framework when adopting other loss/regularization functions, and extend our proposed framework for other applications such as weakly-supervised learning.

\bibliographystyle{IEEEtran}
\bibliography{DSSC}

% Can use something like this to put references on a page
% by themselves when using endfloat and the captionsoff option.
\ifCLASSOPTIONcaptionsoff
  \newpage
\fi

% You can push biographies down or up by placing
% a \vfill before or after them. The appropriate
% use of \vfill depends on what kind of text is
% on the last page and whether or not the columns
% are being equalized.

%\vfill

% Can be used to pull up biographies so that the bottom of the last one
% is flush with the other column.
%\enlargethispage{-5in}

% that's all folks
\end{document}